\documentclass[conference]{IEEEtran}
\IEEEoverridecommandlockouts
\usepackage[misc]{ifsym}

\usepackage{cite}
\usepackage{amsmath,amssymb,amsfonts}
\DeclareMathOperator*{\argmax}{argmax}
\usepackage{graphicx}
\usepackage{textcomp}
\usepackage{xcolor}
\usepackage[breaklinks]{hyperref}
\usepackage[export]{adjustbox}
\usepackage{mathrsfs}
\usepackage{algorithm}
\usepackage{algpseudocode}
\def\BibTeX{{\rm B\kern-.05em{\sc i\kern-.025em b}\kern-.08em
    T\kern-.1667em\lower.7ex\hbox{E}\kern-.125emX}}
\begin{document}

\title{RL-MD: A Novel Reinforcement Learning Approach for DNA Motif Discovery}
\author{
    \IEEEauthorblockN{Wen Wang$^{1}$ , Jianzong Wang$^{1(\textrm{\Letter})}$\thanks{$\textrm{\Letter}$ Corresponding author: Jianzong Wang, jzwang@188.com}, Shijing Si$^{1,2}$, Zhangcheng Huang$^{1}$ and Jing Xiao$^{1}$}
    \IEEEauthorblockA{$^1$Ping An Technology (Shenzhen) Co., Ltd., Shenzhen, China}
    \IEEEauthorblockA{$^2$School of Economics and Finance, Shanghai International Studies University, Shanghai, China}
    \IEEEauthorblockA{Email: wwang.tongji@gmail.com, jzwang@188.com, shijing.si@outlook.com, \\
    huangzhangcheng624@pingan.com.cn, xiaojing661@pingan.com.cn}
}

\maketitle

\begin{abstract}
The extraction of sequence patterns from a collection of functionally linked unlabeled DNA sequences is known as DNA motif discovery, and it is a key task in computational biology. Several deep learning-based techniques have recently been introduced to address this issue. However, these algorithms can not be used in real-world situations because of the need for labeled data. Here, we presented RL-MD, a novel reinforcement learning based approach for DNA motif discovery task. RL-MD takes unlabelled data as input, employs a relative information-based method to evaluate each proposed motif, and utilizes these continuous evaluation results as the reward. The experiments show that RL-MD can identify high-quality motifs in real-world data.
\end{abstract}


\section{Introduction}
\begin{figure}[htbp]
\includegraphics[width=\linewidth]{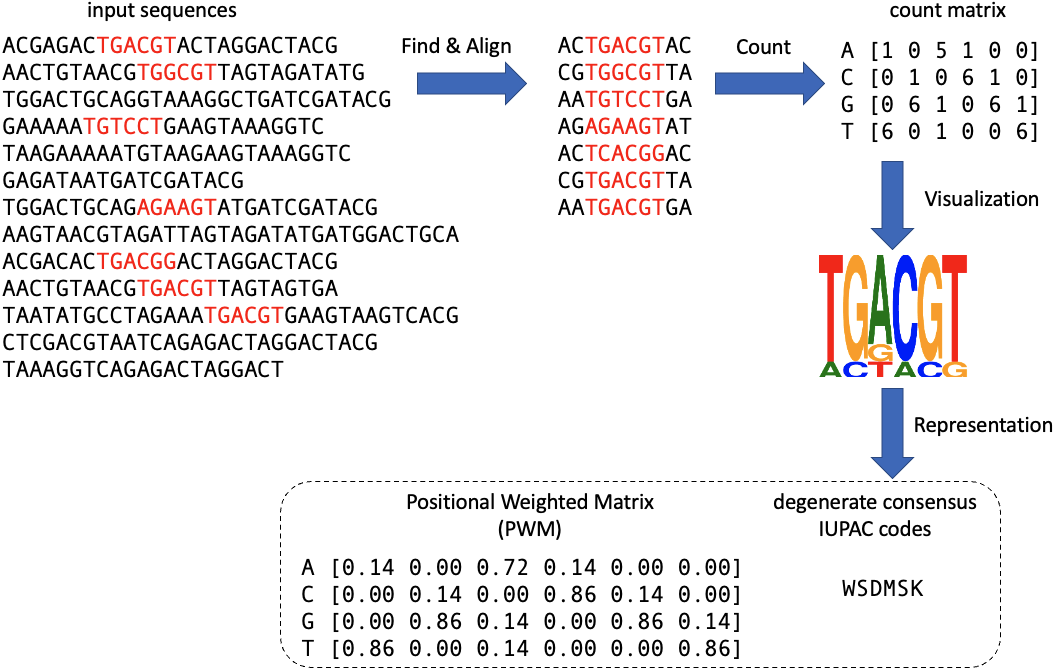}
\caption{The DMD process. DMD receives a set of DNA sequences as input. Some of these sequences have motifs (red text) according to the biological hypothesis. The number of motifs in these sequences is higher than in other types of DNA fragments. DMD is looking for the nucleotide composition of this putative motif. Assuming that the precise location of the motif among the input sequences is known, the motif can be identified by PWM or degenerate consensus IUPAC codes after aligning, counting, and visualizing the motif in a form specifically designed for biologists. In practice, algorithms must perform these operations without position information of the motifs.}
\label{fig.1}
\end{figure}

The DNA sequences that contain the same motif, a specific sequence pattern, are often bound by a particular transcriptional factor (TF) or TF combination. Biologists have shown that TF are crucial in biological processes such as alternative splicing\cite{castle2008expression}, RNA degradation\cite{giraldez2006zebrafish}, and transcriptional regulation\cite{harbison2004transcriptional}. Different types of cells express unique combinations of TFs, which might be viewed as the basic mechanism for cell differentiation\cite{lambert2018human}. Identifying the motif from a collection of unlabeled DNA sequences that hold common regulatory or functional characteristics is a essential task for computational biology known as DNA motif discovery (DMD) (Fig.~\ref{fig.1}). 

The input for the DMD is thousands of sequences, each containing hundreds of nucleotides. Unknown portions of these sequences contain the motif, while the remaining sequences do not. The percentages of sequences containing motifs in different datasets vary due to the poor reproducibility of biological experiments. Input sequences produced by enrichment-based methods like ChIP-seq (chromatin immunoprecipitation followed sequencing)\cite{johnson2007genome} exhibit a dominant motif. Datasets generated by other biological approaches may contain multiple motifs. Here, we focus on the scenario where a dominant motif exists.

Several computational algorithms have been developed to perform DMD. They could be divided into two groups—word-based and profile-based—according to different representation methods for motifs. Word-based approaches utilize (degenerate) consensus IUPAC codes\cite{cornish1985nomenclature} to describe the motif and search the entire input sequence to find the theoretically best motif\cite{bailey2011dreme}\cite{jia2014new}\cite{thomas2012rsat}\cite{ding2014siomics}. The computational complexity increased exponentially when the size of input sequences got a linear increment for these word-based methods. This drawback limits its application in the high-throughput sequencing era. Profile-based techniques represent motifs by a positional weighted matrix (PWM), which can define an infinite number of nucleotide combinations in each position as opposed to a limited number using IUPAC codes\cite{cornish1985nomenclature} (Fig.~\ref{fig.1}). In DMD, profile-based techniques were employed with variable heuristic search technologies\cite{sinha2000statistical}\cite{kulakovskiy2010deep}\cite{ikebata2015repulsive}\cite{kulakovskiy2013binding}. However, the sub-optimal results  generated by these algorithms force users to repeatedly try the DMD process in order to obtain a better motif, which slows down its processing performance.

Several deep learning-based approaches, including DeepBind\cite{alipanahi2015predicting}, basset\cite{kelley2016basset}, HOCNN\cite{zhang2018high}, DeepCpG\cite{angermueller2017deepcpg}, LeNup\cite{zhang2018lenup} and DeFine\cite{wang2018define}, have been introduced to extract sequence features by using the weights in the first convolution layers as the probabilities in PWMs. The training of these discrimination models requires labelled inputs, which distinguish the sequence containing a motif from others. Unfortunately, the existing status of the motif in each sequence in the inputs of DMD is unknown. These limitations make it hard to apply these algorithms in real biological research. Together, a novel unsupervised learning strategy combined with the feature extraction ability of deep learning is needed to improve DMD. Deep reinforcement learning has been considered as a potential approach to address these issues\cite{he2021survey}.

Motivated by the aforementioned limitations, we propose RL-MD, a novel reinforcement learning approach to solve these questions. First, RL-MD is constructed based on DDPG, which does not require a labeled input. Second, RL-MD uses HKDIC (high order Kullback-Leibler divergence), an enhanced motif evaluation approach, to evaluate the motif. Last but not least, RL-MD can produce superior results with data from the real world.

\textbf{Our contributions are presented as follows:}
\begin{itemize}
\item We propose RL-MD to apply reinforcement learning to the task of discovering DNA motifs (DMD). RL-MD find the right DNA motif by the iteration of proposing potential motifs and update its own parameters based on receiving feedback (evaluation score).
\item We design the new evaluation method HKDIC, which takes the adjacent nucleotides combination information into account. We use 2D convolution to conduct this operation. It could extend to a $k$-adjacent nucleotides combination with a low increment in computing time.
\item We test our RL-MD and two other popular algorithms on real-world datasets. In Gcn4 ChIP-seq data, RL-MD was able to obtain a reliable motif, whereas MEME\cite{bailey1994fitting} and ChIPMunk\cite{kulakovskiy2010deep} were unable to produce valid results.
\end{itemize}

\section{Related Works}
\subsection{Biological Background}
DMD is one of the most essential computational biology tasks. Korn\cite{korn1977computer} first introduced a program to analyse the common features among DNA sequences. Following that, researchers introduced algorithms to analyse the motifs in DNA sequences such as the upstream sequence of transcription start sites, transcription termination sites,  and satellite DNA\cite{liu2001bioprospector}.

In the high-throughput sequencing era, ChIP-chip and ChIP-seq generate new types of data for DMD input. Both in terms of number and quality, these inputs are superior to the preceding ones. In quantity, millions of raw signals could be summarized into tens of thousands of peaks which could indicate TF binding sites. In quality, these TF binding sites are supported by physical interaction evidence. The input DNA fragments could be reduced to hundreds base pairs (bp). Previous methods take tremendous time for handle this new data. New algorithms are required for computation biology\cite{lihu2015review}\cite{zambelli2013motif}.

For ChIP-chip data, several algorithms have been developed\cite{liu2001bioprospector}\cite{liu2002algorithm}. Nowadays, the most common input for DMD is ChIP-seq or ChIP-seq-like data. MEME\cite{bailey1994fitting}, the most widely used method, uses a mixture model to describe input sequences and fits this model via expectation maximization to perform DMD. They assume that the input sequences are composed of two sets of sequences. One contains the sequence derived from the motif with random sequences in the flanking regions. The others were formed with random sequences. These two datasets can be described by their model. MEME initiated two models with arbitrary parameters and refined them in the following. In E-steps, MEME calculated the expected value of each sequence derived from models and sent it to the corresponding higher value group. In M-steps, MEME maximizes the model parameters to refine the model. After repeatedly applying these two steps, MEME finally gets the model that can describe the most information about the input sequence. ChIPMunk\cite{kulakovskiy2010deep}\cite{kulakovskiy2013binding} also uses the EM approach. In E-steps, ChIPMunk\cite{kulakovskiy2010deep}\cite{kulakovskiy2013binding} uses the discrete information content (DIC) to measure the similarity between sequences and motif. Its basic version does not consider the background frequency of nucleotides. In an improved version, the discrete version of Kullback–Leibler divergence (Kullback DIC, KDIC) was used, which removed the information content explained by sequence background.

\subsection{Modern Computational Methods}
Convolutional neural network (CNN)-based techniques have been developed recently to address DMD\cite{alipanahi2015predicting}\cite{kelley2016basset}\cite{zhang2018high}\cite{wang2018define}. The inputs for these algorithms are labeled sequences. The labels provide information about whether the corresponding sequence has the target motif. Based on the CNN-based networks, binary classifiers were trained. Benefit from the acceleration of GPU-based tensor calculation, which increases the execution performance of certain methods to a usable level. The weights of the first convolutional layer, which are typically $m$ * 4 in size, were taken into consideration as the PWM of motifs after the training procedure.

Reinforcement learning (RL) has been employed in a variety of tasks for which the ground truths are unknown. The deterministic policy had been introduced for tasks with continuous action space in order to solve these issues utilizing RL\cite{sutton1999policy}. To utilize the benefits of deep learning in feature extraction and deterministic policy, deep deterministic policy gradient (DDPG)\cite{silver2014deterministic} coupled the Actor-Critic technique and deep Q-network (DQN)\cite{mnih2015human}. Prioritized experience replay, which used the priority queue to learn more data on those high bias experiences, has been presented as a way to speed up the training process\cite{hou2017novel}.

\section{Methodology}
\subsection{Applied DDPG in DMD}

\begin{figure}[htbp]
\includegraphics[width=0.5\textwidth]{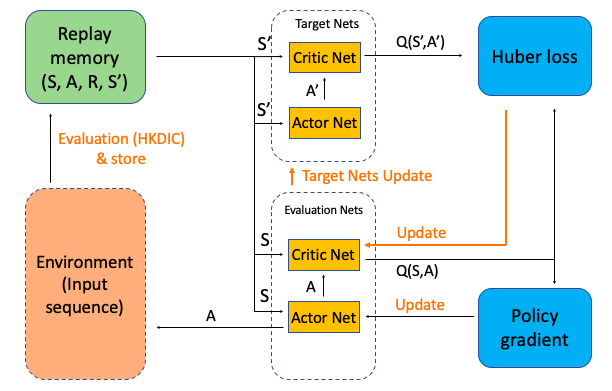}
\caption{The overall structure of RL-MD. RL-MD uses DDPG as the general framework. The actor network $\mu$ first generate some random actions and received feedback (reward) from the environment. Replay memory $D$ was used to store these states, actions, rewards, and new states. The actor network was updated according to the policy gradient. The critic network $Q$ was updated according to the difference between the evaluated value of an action given by the critic network and predicted value using the target network and reward.}
\label{fig.2}
\end{figure}

The purpose of the RL agent is to maximize the total reward $R = \displaystyle \sum_{t=t_0}^{\infty} \gamma^{t-t_0} r_t$. Theoretically, there is an optimal policy $\pi^*$ that can guide our choice of action $a \in \mathscr{A}$ for each state $s \in \mathscr{S}$ in order to accomplish this goal. The evaluation function $Q$ that describes the reward $R$ in each state $s$ and action $a$ combination is need.

\begin{equation} \label{equ_1}
    Q(s,a) \to R
\end{equation}
The optimal policy $\pi^*$ could be obtained by maximizing the rewards received according to $Q$.
\begin{equation} \label{equ_2}
    \pi^*(s) = \argmax_a Q(s,a)
\end{equation}
Given that the action space in RL-MD is continuous, deep learning networks have been employed to approximate the evaluation function $Q$ as a $Q^{\pi}$. With reference to the bellmen equation, $Q^{\pi}$ is updated.
\begin{equation} \label{equ_3}
    Q^{\pi}(s, a) = r + \gamma Q^{\pi}(s', \pi(s'))
\end{equation}
Temporal difference error (TD error) $\delta$ describes the difference between the two sides of the equality:
\begin{equation} \label{equ_4}
    \delta = Q(s, a) - (r + \gamma \max_a Q(s', a))
\end{equation}
$Q^{\pi}$ will eventually approach $Q$ by updating the network using the loss at each step (Fig.~\ref{fig.2}). Taken together, we proposed the RL-MD in algorithm \ref{alg}.

\subsubsection{Action}
In RL-MD, the actor networks will generate potential motifs as the actions. The motif is a probability matrix with m * 4 sizes, denoted as $M$ here, and shown in Fig.~\ref{fig.1}. $M_{i,j}$ is the probability of the $j$-th nucleotide at the $i$-th position.

\subsubsection{State}
The environment of RL-MD is the input sequences, a set of unlabeled DNA sequences. The state $S$ of the environment is the detailed nucleotides for each input sequence and stays constant throughout the whole process. The state $S$ are loaded into RL-MD at the beginning rather than being saved in the replay memory. Biological experiments produced DNA fragments of varying lengths for DMD input. We clipped the edges on both sides to reserve the 100 bp around the signal summits for convenience.

\subsubsection{Reward}
Similar to ChIPMunk\cite{kulakovskiy2010deep}\cite{kulakovskiy2013binding}, RL-MD evaluated how a motif could explain the information in input sequences. These measurement are used by RL-MD as the reward $R$ in RL. For the sake of computational simplicity, previous algorithms\cite{kulakovskiy2010deep}\cite{kulakovskiy2013binding} assumed that the distribution of nucleotides between adjacent DNA sites was independent. However, biologists have revealed that the relationship between adjacent nucleotides may be essential for some biological phenomena, such as the under-expectation frequency of the cytosine-phosphate-guanine dinucleotide (CpG)\cite{sved1990expected}. The bias of DNase-seq data was discovered by methods that took into account the relationship between high-order neighboring nucleotides\cite{he2014refined}. As a substitute for KDIC, RL-MD uses HKDIC. The original KDIC could be regarded as zero-order HKDIC, which does not consider adjacent nucleotides. In the first-order HKDIC, we consider the first adjacent nucleotide, so the type of nucleotides in each position is 64 ($4^{2 \times 1 + 1}$) rather than 4 ($4^{0 \times 1 + 1}$). The order of HKDIC is a hyper-parameter in RL-MD and is set as 1 in the following.
The background frequency for each nucleotide were denoted as:
\begin{equation} \label{equ_5}
    Q = (q_1, q_2, \cdots, q_k)
\end{equation}
where $k$ is the number of different types of nulcleotides in the input sequences. For the first-order HKDIC, $k=64$.

Suppose that there are $N$ DNA sequences contained in the input sequences, and that the composition of nucleotides at a specific position is :
\begin{equation} \label{equ_6}
    C = (c_1, c_2, \cdots, c_k)
\end{equation}
where $\displaystyle\sum_{i=1}^k c_i = N$. The likelihood of this occurrence is:
\begin{equation} \label{equ_7}
\begin{split}
    P(q_1, q_2, \cdots, q_k | Q) = \\
    \frac{N!}{c_1 \times c_2 \times \cdots \times c_k} q_1^{c_1} \times q_2^{c_2} \times \cdots \times q_k^{c_k}
\end{split}
\end{equation}
After the logarithm:
\begin{equation} \label{equ_8}
\begin{split}
    log P(q_1, q_2, \cdots, q_k | Q) = \\
    log \frac{N!}{c_1 \times c_2 \times \cdots \times c_k} + 
    \displaystyle\sum_{i=1}^k c_i q_i
\end{split}
\end{equation}
The KDIC is defined as:
\begin{equation} \label{equ_9}
\begin{aligned}
    KDIC & = \frac{1}{N}(\displaystyle\sum_{i=1}^k log x_i ! - log N !) - \displaystyle\sum_{i=1}^k \frac{x_i}{N} log q_i
\end{aligned}
\end{equation}
where $x_i$ is the proportion of $i$-th nucleotide in this position.
In HKDIC, the number of nucleotide types $k$ is different according to the order $o$ taken into consideration. Thus, the HKDIC for a single position is here:
\begin{equation} \label{equ_10}
    HKDIC = \frac{1}{N}(\displaystyle\sum_{i=1}^{k^{2 \times o+1}} log x_i ! - log N !) - \displaystyle\sum_{i=1}^{k^{2 \times o+1}} \frac{x_i}{N} log q_i
\end{equation}

To calculate the reward for a given motif $a$ to the input sequences $\mathscr{S}$, an optimal alignment set $P$ will be generated first. The optimal alignment between each sequence $s_i \in \mathscr{S}$ and motif matrix $M$ will be calculated in both strands of the DNA sequence. The DNA fragment aligned to the motif matrix of the better result in these two alignment will be saved as $p_i$. The $p_i$ is a $k \times m$ sized matrix, where $k$ is the number of nucleotides and $m$ is the length of motif.
The value in $p_i$ is defined as:
\begin{equation} \label{equ_11}
    p_{i,k,m} = \begin{cases}
    1, \quad \text{the $m$-th position is the $k$-th nucleotide} \\ 
    0, \quad \text{otherwise}
    \end{cases}
\end{equation}
The $P$ is the sum of each $p_i$, $P = \displaystyle\sum_{i=1}^N p_i$. And the reward $R$ is sum of HKDIC in each position:
\begin{equation} \label{equ_12}
    R = \frac{1}{N}(\displaystyle\sum log x_{i,m} ! - log N !) - \displaystyle\sum \frac{x_{i,m}}{N} log q_i
\end{equation}
where $x_{i,m}$ is the count of $i$-th nucleotide in the $m$-th position in $P$, $q_i$ is the frequency of the $i$-th nucleotide in the input sequences.

\subsection{Network Structure}

\begin{figure}[htbp]
\includegraphics[width=0.5\textwidth]{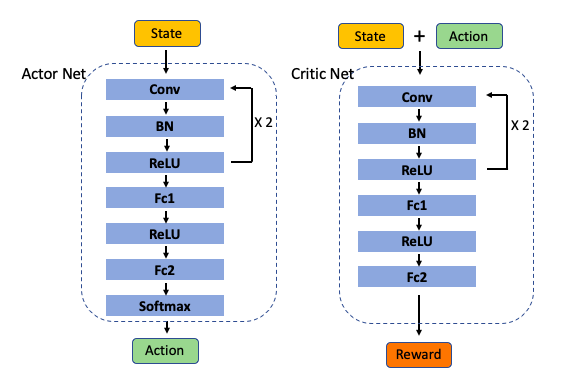}
\caption{Network structure in RL-MD. Actor networks take the state $s$ (input sequences of DMD) as input and output the proposed motif as action $a$. Critic networks take the states $s$ and actions $a$ generated by actor networks as input, then output the reward $R$.}
\label{fig.3}
\end{figure}

The state $\mathscr{S}$ input to actor networks is a $2 \times N \times 100 \times 4 $ sized tensor. Actor networks have two blocks of convolution layers, batch normal layers, and activation layers. Actor networks contain two fully connected layers with softmax function in the following and output a $m \times 4$ sized matrix action $a$.
A $2 \times N \times (100 + m) \times 4 $ sized tensor, which contains both the state $\mathscr{S}$ and action $a$ information, is the input of critic networks. Critic networks have two blocks of convolution layers, batch normal layers, and activation layers. Actor networks contain two fully connected layers in the following and output an evaluation score about the input action $a$ to state $\mathscr{S}$ (Fig.~\ref{fig.3}).

\subsection{Exploration Enhancement}
RL-MD uses the soft-$\epsilon$ policy to improve the exploration process in order to solve the local optimal problem. Typically, during the exploration process, motifs (action $a$) were produced at random. For most positions in these generated motifs, the frequency of each nucleotide is similar. However, in most positions in real-world motifs, the nucleotide frequency is extremely unbalanced. This difference in nucleotide distribution between motifs that are generated at random and those in the real world causes a reduction in exploration efficiency. RL-MD addresses this issue by adopting rejection sampling. The inequality index $e$ of the whole motif is defined as:
\begin{equation} \label{equ_13}
    e = \displaystyle\prod_{i=1}^{m} e_i
\end{equation}
where $i$ is the position index of the motif. The inequality index $e_i$ for each position in the motif is defined as:
\begin{equation} \label{equ_14}
    e_i = (\frac{k \displaystyle\sum_{i=1}^k \displaystyle\sum_{j=1}^k | m_i - m_j |}{2 n^2})^4
\end{equation}
The greater the nucleotide frequency inequality inside the provided motif, the lower the inequality value will be and the greater the likelihood that this motif will pass the rejection sampling.


\subsection{Weighted Experience Replay}
The main objective of the critic network is to assist the actor network in taking better action with higher reward. Therefore, the action that offers a greater reward is more important for the networks to learn. To focus on these high-value occurrences, RL-MD employs a weighted experience replay strategy. Weighted sampling is used to construct the learning batches when the target networks retrieve the memory, and the reward values are employed as the weights.

\section{Experiments}

\begin{algorithm}
\caption{RL-MD’s algorithm}\label{alg}
\begin{algorithmic}[1]
\State Read the input sequences and calculate the count each type of nucleotides  \Comment{The types of nucleotides are determined by the order of HKDIC $o$}
\State Randomly initialize the actor network and critic network in evaluation networks with parameters $Q(s,a|\Theta^Q)$ and $\mu(s|\Theta^\mu)$
\State Initialize the actor network and critic network in target networks with parameters $\Theta^{Q'} \gets \Theta^Q$ and $\Theta^{\mu'} \gets \Theta^\mu$
\State Initialize the replay memory  \Comment{This is a heap with fix length $D$}
\For {$episode = 1, E$}
    \If {$E <= 100$}
        \For {$t = 1, T$}
            \State Generate a random action
            \State Get the reward from the environment
            \State Save this action-reword pair into replay memory
        \EndFor
    \Else
        \State Initialize the adaptive Ornstein-Uhlenbeck process
        \For {$t = 1, T$}
            \State $a \gets A \sim \mathcal{U}(0,1)$
            \If {$a <= \epsilon$}
                \State Generate a random action
            \Else
                \State Generate a action according to the actor network in evaluation networks
            \EndIf
            \State Get the reward from the environment
            \State Save this action-reword pair into replay memory
            \If {$episode \times (T - 1) + t > D$}
                \State Weighted sampling a random minibatch from the replay memory
                \State Set $y_i = r_i + \gamma Q'(s_{i+1},\mu'(S_{i+1}|\Theta^{\mu'})|\Theta^{Q'})$
                \State Update the critic networks by minimizing the huber loss
                \State Update the actor networks according to the policy gradient
                \State Update the target networks: \\
                    \quad \quad \quad \quad \quad \quad $\Theta^{Q'} \gets \tau \Theta^{Q} + (1 - \tau) \Theta^{Q'}$ \\
                    \quad \quad \quad \quad \quad \quad $\Theta^{\mu'} \gets \tau \Theta^{\mu} + (1 - \tau) \Theta^{\mu'}$
            \EndIf
        \EndFor
    \EndIf
\EndFor
\end{algorithmic}
\end{algorithm}

\subsection{Experimental Setup}
The evaluation was carried out using the CTCF and Gcn4 motifs. There are two different artificial dataset types and one real-world dataset used in the evaluation. The first type of artificial dataset is a collection of all positive (with motif) DNA sequences. The second is a mixture of DNA sequences that are two thirds negative (without a motif) and one third positive. Data from ChIP-seq is used to create the real-world dataset. Full-pos, unbalanced, and ChIP-seq labels were assigned to these datasets individually.

\begin{figure}
\includegraphics[width=0.5\textwidth]{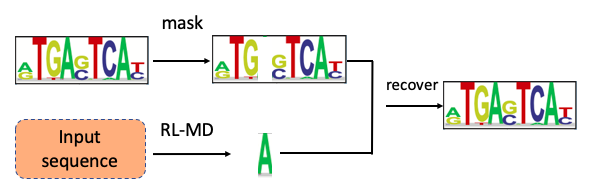}
\caption{Test for single-nucleotide recovery. In each individual test, a position was randomly selected and masked in the motif. The output shapes of the actor networks were set to $1 \times k$ to fill in this gap. The synthesised motifs were used in the following process. These tests could access the ability of RL-MD to provide the right nucleotide frequency at a single nucleotide position.}
\label{fig.4}
\end{figure}

\begin{figure*}[htbp]
\includegraphics[width=\textwidth]{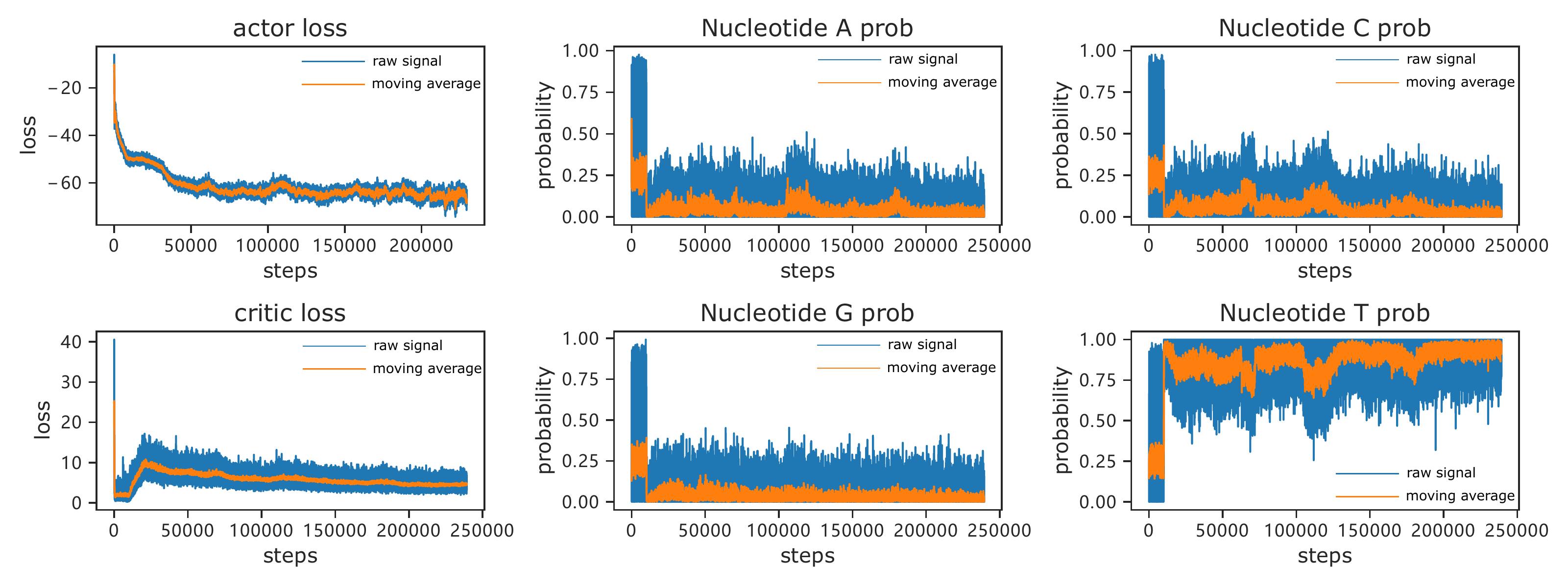}
\caption{Example of a training process. The two figures on the left side describe the training loss patterns for the actor network (upper panel) and critic network (lower panel). The probabilities of four nucleotides in the proposed motifs by the actor network are shown in the middle and right figures. These results suggest that throughout the training phase, both actor and critic networks can converge. The navy and orange lines show the raw and 50-length moving average signals, respectively.}
\label{fig.5}
\end{figure*}

\begin{table*}\centering
\caption{E-value measurement of output motif compared with ground truth}
\begin{tabular}{ c c c c c c c }
\hline
\multicolumn{1}{c}{} & \multicolumn{3}{c}{Gcn4} & \multicolumn{3}{c}{CTCF} \\
 & full-pos & imbalanced & ChIP-seq & full-pos & balanced  & ChIP-seq \\
\hline
MEME\cite{bailey1994fitting} & \textbf{2.7e-07} &  \textbf{3.2e-06} & N.A. & 9.1e-28 &  \textbf{1.3e-40} & 4.6e-18 \\
ChIPMunk\cite{kulakovskiy2010deep} & 8.0e-06 & 2.3e-03 & N.A. & \textbf{0} & 8.6e-36 & \textbf{1.2e-31} \\
RL-MD & 2.0e-04 & 9.4e-05 & \textbf{6.6e-02} & 6.0e-06 & 5.9e-10 & 5.1e-07 \\
\hline
\end{tabular}
\label{table.1}
\end{table*}

The motifs discovered with different algorithms were saved in PWM format. To assess the similarity between two motifs, several assessment criteria have been developed\cite{pietrokovski1996searching}\cite{choi2004local}. Here, E-value is used to perform this measurement\cite{gupta2007quantifying}. The more similar the detected motif and ground truth are, the lower the E-value will be. TOMTOM\cite{gupta2007quantifying} was used to calculate this similarity between motifs discovered by different algorithms and motifs from the JASPAR database (JASPAR 2022 core non-redundant)\cite{castro2022jaspar}.

\subsubsection{Single nucleotide rescue}
A simplified method is utilized to assess the capacity of this algorithm before performing the DMD using RL-MD. A position among known motifs will be masked. The algorithm should recover this position using the full-pos datasets derived from this motif. A $1 \times k$ sized matrix that is appropriate for this position is what RL-MD seeks to deliver. The output of actor networks will be used to complete the original masked position in the motif (Fig.~\ref{fig.4}). This entire motif will be used for both the evaluation of the environment and input for critic networks.
The 13-th position of the Gcn4 motif was covered. Within 10,000 steps, RL-MD can retrieve it (Fig.~\ref{fig.5}). All positions in the CTCF motif as well as other positions in Gcn4 motif were tested using this method. RL-MD was able to pass the test in each positions of these two motifs. These findings demonstrate that RL-MD was able to identify the proper nucleotide frequency composition in the motif.

\subsubsection{Artificial Data}
The artificial datasets were created in the following steps. First, the target motifs (CTCF and Gcn4) were obtained from the JASPAR database\cite{castro2022jaspar}. For the positive sequences (with motif), the central sequence was generated using the PWM of motif, and the flanking sequences were added using random nucleotides to create a 100 bp-length sequence. The negative sequences were made up entirely of random nucleotides. Three hundred positive sequences were used for the full-pos dataset. The unbalanced datasets have 200 negative and 100 positive sequences.

\subsubsection{Real World Data}
The Gcn4 and CTCF ChIP-seq data were downloaded from Gene Expression Omnibus (GEO)\cite{barrett2012ncbi} under GEO accession numbers GSE85588\cite{mittal2017gcn4} and GSE30263\cite{wang2012widespread}, respectively. The ChIP-seq peaks were sorted by the enrichment signal and preserve the top 300. The lengths of peaks were extended to 100 bp from the peak summits using BEDtools\cite{quinlan2010bedtools}. The input sequences were generated according to the normalized peaks and genome sequence.

\subsubsection{Algorithm and hyper-parameters}
We set the hyper-parameters in RL-MD as: the maximum episode $E = 10000$, the steps in each episode $T = 10$, the capacity of the replay memory $D = 1000$, the exploration control parameter $\epsilon = 0.01$, the target networks update ratio $\tau = 0.1$, the order of HKDIC $o = 1$, and the learning rate for actor and critic networks $\gamma$ were selected from $\{1 \times 10^{-2},1 \times 10^{-3},1 \times 10^{-4},1 \times 10^{-5}\}$ to find the best results.

\subsection{Baseline}
Two commonly utilized algorithms, MEME\cite{bailey1994fitting} and ChIPMunk\cite{kulakovskiy2010deep}, were used using actual biological data. Each year, the MEME was cited more than 200 times, and in the last five years, ChIPMunk has been cited 50 times. TOMTOM\cite{gupta2007quantifying} were used to assess the output motif by these two algorithms. In order to shorten the running time of MEME, the number of output motifs was changed from the default value of 20 to 5.

\subsection{Results}

\subsubsection{CTCF}
CTCF is a TF that is found throughout many species and is important for high-order chromatin structure\cite{tang2015ctcf}. A highly conserved motif with a spacing-CCCG pattern marks the binding sites of CTCF. In comparisons of DMD approaches, the DNA motif discovery process for CTCF was typically used. Here, CTCF was used to compare RL-MD, MEME, and ChIPMunk. A good DMD in biology is defined as an E-value between output motif and ground truth that is less than $1 \times 10^{-3}$. In the final evaluation process, all three algorithms can provide good results on the two artificial datasets. Besides, RL-MD also produced CTCF motif results that were comparable to those of the MEME and ChIPMunk using real-world data (Table.~\ref{table.1}).

\subsubsection{Gcn4}
Gcn4 is a TF with a well-defined variable spacing motif\cite{siggers2014protein}. All three approaches produced reliable motifs in simulated data, however the E-value is much greater than CTCF. These results indicate that the DMD on Gcn4 is a challenging task. The MEME and ChIPMunk both failed in the task using the real-world data. The N.A. indicates that the results generated by algorithms do not satisfy the need for TOMTOM to provide an E-value (Table.~\ref{table.1}).

\section{Conclusions}
In this paper, we proposed a novel reinforcement learning-based computational framework, RL-MD, for DMD. RL-MD could take the input sequences and output the required motifs. In real-world data, RL-MD produces a better result for Gcn4. However, we are unable to achieve better results in the simulation dataset. These results may resulted from the lack of attention on the motifs that have already been recognized during the exploration process. Future research may focus on improving the noise generation process so that the agent can test additional potential motifs close to the identified motifs with high evaluation feedback.
Besides, the more complicated network structures may be used in the future to improve the feature extraction ability and evaluate the final results.

\section{Acknowledgement}
We thank X. Qu and N. Cheng for advice and suggestions. This paper is supported by the Key Research and Development Program of Guangdong Province under grant No.2021B0101400003. Corresponding author is Jianzong Wang from Ping An Technology (Shenzhen) Co., Ltd (jzwang@188.com).

\newpage

\bibliographystyle{IEEEtran}
\bibliography{mybib}

\end{document}